\documentclass[letterpaper, 10 pt, conference]{ieeeconf} 
\IEEEoverridecommandlockouts
\usepackage{siunitx}
\usepackage{physics}
\usepackage[T1]{fontenc}
\AtBeginDocument{\RenewCommandCopy\qty\SI}
\usepackage{amsmath,amsfonts}
\usepackage{array}
\usepackage{cite}
\usepackage[caption=false,font=scriptsize]{subfig}
\usepackage{textcomp}
\usepackage{stfloats}
\usepackage{float}
\usepackage{url}
\usepackage{verbatim}
\usepackage{graphicx}
\usepackage{amssymb}
\usepackage{pifont}
\usepackage{algorithm}
\usepackage{algpseudocode}
\usepackage{todonotes} 
\usepackage[hidelinks]{hyperref}
\usepackage[none]{hyphenat}
\usepackage[capitalize]{cleveref}
\usepackage{comment}
\usepackage{multirow}
\usepackage{booktabs}
\usepackage{rotating}
\usepackage{makecell}
\usepackage{tabularx}
\usepackage{xcolor}
\usepackage{blindtext}
\usepackage{algorithm}
\usepackage{algpseudocode}
\usepackage{pgf}
\usepackage[utf8]{inputenc}
\usepackage[T1]{fontenc}
\usepackage[inkscapelatex=false, inkscapepath=./build/svg-inkscape]{svg}

\def\BibTeX{{\rm B\kern-.05em{\sc i\kern-.025em b}\kern-.08em
    T\kern-.1667em\lower.7ex\hbox{E}\kern-.125emX}}
    
\begin{document}

\title{\LARGE \bf
Talk to Me, Jarvis: An Open-Source Edge-Deployable Voice Assistant Framework for Autonomous Racecars
}


\author{Daniel Henel, Frederik Werner, Alexander Langmann, Johannes Betz %
\thanks{D. Henel, A. Langmann and J. Betz are with the Professorship of Autonomous Vehicle Systems, TUM School of Engineering and Design, Technical University of Munich, 85748 Garching, Germany; Munich Institute of Robotics and Machine Intelligence (MIRMI)}
\thanks{F. Werner is with the Institute of Automotive Technology, TUM School of Engineering and Design, Technical University of Munich, 85748 Garching, Germany; Munich Institute of Robotics and Machine Intelligence (MIRMI).
This work has been accepted for publication in the Proceedings of the IEEE Intelligent Transportation Systems Conference (ITSC 2026)}
}





\maketitle


\begin{abstract}
Recent advances in large language models have improved their effectiveness as back-end components for voice assistants, particularly in intent understanding and context-aware input classification. However, online-hosted models introduce network dependency and variable inference latency, limiting their suitability for time-critical autonomous driving applications.

In this work, we address these issues by developing Jarvis, an offline voice assistant for high-level behavioral commands of autonomous vehicles. Its architecture integrates speech recognition and synthesis with natural language command classification into a lightweight, local framework.

Jarvis' core component is a text-to-command classifier, built using a domain-specific fine-tuning of the Mistral 7B model, demonstrating low-latency inference. Our experimental evaluation demonstrates that our solution outperforms larger online-hosted models, achieving 97.63\,\% intent recognition accuracy with an average processing latency of 1.39\,s, making it well-suited for operations requiring quick response times. To support further research and fine-tuning, we provide an open-source implementation.

\textit{Index Terms—}voice assistant, autonomous driving, intent recognition, large language model

\end{abstract}

\section{INTRODUCTION}
In racing, fractions of a second can decide between victory and defeat. Autonomous racing vehicles must make split-second decisions to outperform competitors, placing high demands on their software systems \cite{Betz.2022}. This requirement extends to the AI models that support these vehicles, which must be optimized to ensure both speed and efficiency while maintaining reliable performance.

Despite a high degree of autonomy, human input is still needed in certain situations in autonomous racing, e.g., issuing high-level commands like start or stop, or reacting to unexpected situations on the track that require sending the car to the pit lane. Traditional interfaces for these commands are usually provided through a graphical user interface (GUI) or a command line, which require visual attention and increase reaction time.

To address this problem, voice assistants provide an alternative solution, allowing the human operator to issue commands without looking away from the live telemetry. However, developing such voice assistants is not straightforward. Challenges like latency, command detection reliability, and network dependency need to be addressed to design a useful voice assistant system. 

Another key challenge arises from the inherent mismatch between the unbounded space of natural language expressions and the limited set of executable commands. For example, an operator may issue a \textit{“stop”} command in many different ways - \textit{“halt the car”}, \textit{“bring it to a stop”}, or simply \textit{“stop”} - all of which must be correctly recognized and mapped to the same underlying action. This variability motivates the use of Large-Language-Models (LLMs) to capture semantic equivalence across phrasings.

\subsection{Contributions}
This work makes the following contributions:
\begin{itemize}
\item A comparison of online-hosted and locally deployed language models for voice command classification, evaluating both inference latency and intent recognition accuracy.
\item A fine-tuning and deployment pipeline for a lightweight local model that achieves superior intent recognition accuracy while reducing inference latency compared to online alternatives.
\item An open-source, offline voice assistant architecture integrating wake-word detection, speech transcription, and command classification, demonstrated on resource-constrained mobile hardware (\href{https://github.com/TUM-AVS/jarvis_voice_assistant}{\texttt{\nolinkurl{github.com/TUM-AVS/jarvis_voice_assistant}}}).
\end{itemize}

\begin{figure}[t]
    \centering
    \includegraphics[width=\columnwidth]{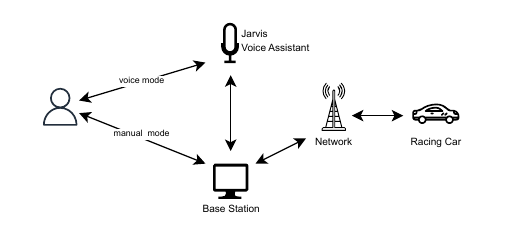}
    \caption{System Architecture: Jarvis is a voice assistant for human operators used to control a autonomous racecar via voice commands.}
    \label{fig:system_architecture}
\end{figure}

\begin{figure*}[!t]  
    \centering
    \includegraphics[width=\textwidth]{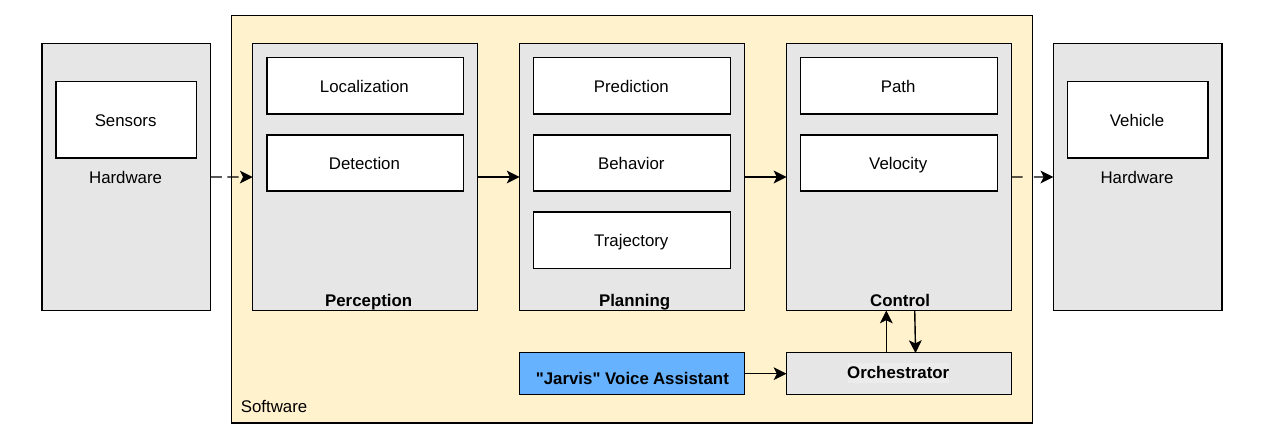} 
    \caption{Autonomous driving software modules.}
    \label{fig:system_components}
\end{figure*}

\subsection{Related Work}

LLMs are increasingly used in various components of autonomous driving, such as scene understanding and multimodal perception~\cite{liu2023llava,cui2024survey}, planning and decision making~\cite{mao2023gptdriver, sha2023languagempc}, explainability and human-machine interaction~\cite{xu2023drivegpt4, sima2024drivelm}, commonsense and traffic rule reasoning~\cite{chen2024drivingwithllms, li2024drivingeverywhere}.

Many researchers investigate the use of LLMs in intent recognition and command classification~\cite{Yang_2024_WACV, pmlr-v235-he24a, Seegert_2026_Modular_Autonomy}, focusing on translating natural language or contextual cues into actionable driving behaviors. Cui et al.~\cite{cui2025llm4adlargelanguagemodels} propose LLMs for Autonomous Driving (LLM4AD), which conceptualize the LLM as a cognitive brain for autonomous vehicles, capable of reasoning and intent understanding. By leveraging the broad knowledge of large-scale models like GPT-4~\cite{openai2024gpt4}, LLM4AD aims to interpret abstract human inputs, such as adjusting driving style e.g. because a passenger is in a hurry, and translate them into high-level control strategies. However, while LLM4AD provides a benchmark for general passenger comfort, it relies on cloud-based Application Programming Interfaces (APIs) that introduce notable inference latencies and depend on a stable network connection. Furthermore, the framework is primarily designed for civilian vehicles with occupants. In contrast, our approach targets autonomous racing, shifting the focus to the interface between a race engineer and the autonomous driving system operating the vehicle. This scope requires a responsive architecture that is more focused on accurate command understanding and low latency and less on general human-centric interaction.

Domain-specific adaptation of lightweight Small Language Models (SLMs)~\cite{lu2025smalllanguagemodelssurvey} has proven effective across a variety of specialized tasks~\cite{gu2026minillmonpolicydistillationlarge, hsieh2023distillingstepbystepoutperforminglarger, WANG2026104035}. For instance, Pal et al.~\cite{pal2025aipowered}  demonstrated that supervised fine-tuning combined with Quantized Low-Rank Adaptation (QLoRA)~\cite{dettmers2023qloraefficientfinetuningquantized} on the Phi-3.5 Mini Instruct model~\cite{abdin2024phi3} significantly improves intent recognition in vehicle-related tasks. This architecture addresses critical challenges such as computational constraints and latency by employing 4-bit quantization to enable efficient execution on edge hardware. Our work adopts a similar technical approach, described in detail in Section~\ref{subsec:fine_tuning}.

Beyond LLM-based intent recognition, the operational viability of voice-enabled assistants in time-critical applications is fundamentally tied to the underlying speech-to-text (STT) architecture~\cite{Rista202086112, nayeem2025automaticspeechrecognitionmodern}. Di Leo et al.~\cite{dileo2025realtime} propose a real-time STT framework specifically designed for edge computing environments requiring ultra-low latency and local execution. Their prototype utilizes the lightweight Vosk engine~\cite{soni2025improvingspeechrecognitionaccuracy} for offline automatic speech recognition (ASR)~\cite{nayeem2025automaticspeechrecognitionmodern}, focusing on data privacy and operational independence by eliminating reliance on external network bandwidth. The authors highlight that their modular and extensible design is intended to accommodate the future integration of more advanced neural STT engines, such as OpenAI's Whisper~\cite{radford2023whisper}. Our work adopts this local-first philosophy but directly utilizes the Whisper architecture, as described in detail in Section~\ref{subsec:jarvis_architecture}.


\section{Method}\label{sec:method}
We integrate our \textit{Jarvis} voice assistant with a software stack for autonomous motorsport applications \cite{{Hoffmann.2026}}. We describe the integration in Section \ref{subsec:system_architecture} and the \textit{Jarvis} architecture itself in Section \ref{subsec:jarvis_architecture}.

\subsection{System Architecture}\label{subsec:system_architecture}

The system is centered around a base station (Figure~\ref{fig:system_architecture}), which acts as the interface between humans and the onboard vehicle software. Moreover, it receives information about the status of the racing car and all associated software modules (Figure~\ref{fig:system_components}), providing telemetry, and diagnostics. The software modules include, among others, perception, planning, detection, which communicate via the Robot Operating System 2 (ROS2) \cite{quigley2009ros}. Our \textit{Jarvis} voice assistant is implemented as a stand-alone module and extends the base station by enabling a human operator to issue high-level commands using natural speech. Speech inputs are processed by the voice assistant pipeline, as described in Section~\ref{subsec:jarvis_architecture}, where automatic speech recognition and intent interpretation generate predefined behavioral commands, such as speed targets, start/stop commands or requests to exit or return to the pit lane. These commands are transmitted to the base station using ROS2, where they undergo validation and safety checks before being accepted. Once validated, the base station forwards the commands to the racing car over a mobile network connection to the vehicle. To ensure transparency and operator awareness, every voice-enabled command is visualized in the base station interface, allowing the human controller to verify correct command detection and execution.

\subsection{Voice Assistant Architecture}\label{subsec:jarvis_architecture}

\begin{figure*}[!t]  
    \centering
    \includegraphics[width=\textwidth]{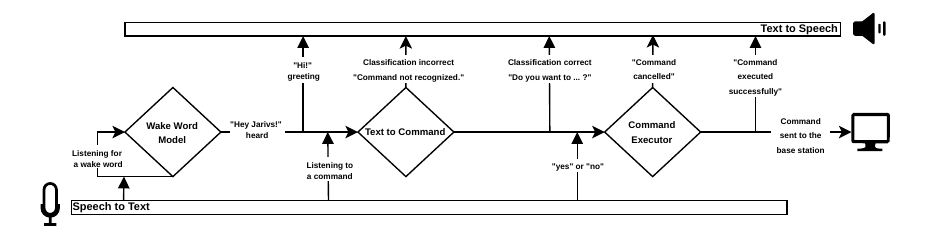} 
    \caption{System architecture and process flow of Jarvis. After wake-word detection, the system performs local speech-to-text transcription, LLM-based command classification, and text-to-speech feedback. Confirmed commands are transmitted via ROS2 to the base station, where safety checks precede transmission on the autonomous racing vehicle.}
    \label{fig:jarvis_architecture}
\end{figure*}

The architecture of \textit{Jarvis} is visualized in Figure~\ref{fig:jarvis_architecture}. The system is initiated by wake-word detection based on the openWakeWord framework\footnote{\url{https://github.com/dscripka/openWakeWord}}, which continuously monitors the audio stream for the trigger phrase “Hey Jarvis!”, using keyword spotting techniques~\cite{chen2014smallfootprint}. Upon activation, the assistant acknowledges the human operator with a short synthesized response "Hi!" and transitions into a command-listening state.

Speech-to-text conversion is performed locally using OpenAI’s Whisper model, specifically the lightweight English-only “base.en” variant\footnote{\url{https://github.com/openai/whisper}}, enabling low-latency, network-independent transcription. The resulting text is forwarded to the core LLM-based text-to-command classifier, which interprets the operator’s intent and maps the natural-language input to a predefined behavioral command. We compared online-hosted and locally deployed open-weight models for command classification. While cloud models provide strong zero-shot performance, network-based inference introduces higher latency. Fine-tuned local models achieved the best trade-off between latency and accuracy as described in Section~\ref{sec:results}.

Following command interpretation, auditory feedback is generated for the operator using a text-to-speech (TTS) module based on the Coqui TTS framework\footnote{\url{https://github.com/coqui-ai/TTS}} and a VITS-based neural speech synthesis model~\cite{kim2021vits}. This feedback confirms successful command recognition or indicates failure. Frequently used responses are pre-recorded to reduce latency and played back directly.

In the final step, the human operator is asked to confirm the execution of a recognized command, and only confirmed commands are transmitted to the base station via ROS2, where additional safety checks and state validations are performed before forwarding the commands to the autonomous racing car.

\section{RESULTS}
\label{sec:results}

We conducted a multi-stage evaluation. First, we compared multiple LLMs in default configurations. We then applied two successive fine-tuning stages, resulting in the final selected model. Our primary benchmarking criteria were:
\begin{itemize}
    \item \textbf{Intent Classification Accuracy} defined as the ratio of correctly classified commands to the total number of test samples.
    \item \textbf{Inference time} measured as the end-to-end processing time required for the model to classify a command. The measurement interval began the moment the tokenized input was fed into the model and ended once the final "end-of-sequence" token was generated. 
\end{itemize}

\subsection{Initial benchmarking of LLMs}
\label{subsec:initial_benchmarking}

We first report benchmark results for multiple LLMs evaluated without task-specific fine-tuning comprising of several online-hosted and locally deployed models in terms of inference time and initial classification accuracy. The benchmark was conducted using an expert-curated dataset of predefined high-level racing commands. The dataset of 85 samples was designed for preliminary evaluation and was paired with a prompt that described all commands and enforced a structured output format to ensure consistent command interpretation.

We evaluated OpenAI GPT-4, GPT-3.5, and o3-mini variants, listed in Table \ref{tab:model_benchmarks}. All models were accessed via the OpenAI \footnote{\url{https://developers.openai.com/api/reference/overview/}} API under a paid account plan, using default settings, including the temperature equal 1.0.

Additionally, we benchmarked locally deployed, open-weight SLMs from the Gamma, DeepSeek, Llama, and Mistral families. These experiments were run on an NVIDIA GeForce RTX 4090 GPU.

The results, summarized in Table~\ref{tab:model_benchmarks}, reveal a trade-off between accuracy and latency. Online models generally achieved higher classification accuracy but often exhibited prohibitively long response times, making them unsuitable for our target application. In contrast, locally deployed models showed lower accuracy but significantly reduced inference latency.
Given the quick response time demand of the target application, latency was prioritized over accuracy. Consequently, \texttt{Mistral 7B}~\cite{jiang2023mistral7b}, \texttt{LLaMA-3.1 8B}, and \texttt{LLaMA-3.2 3B}~\cite{grattafiori2024llama3herd}, the three models with the lowest prompt processing times, were selected for further fine-tuning. Initial accuracy deficiencies were considered secondary, as classification performance was expected to improve through fine-tuning on a domain-specific dataset.

\begin{table}[t]
\centering
\caption{Initial benchmarking of online and local Models. Highlighted models were selected for further fine-tuning.}
\label{tab:model_benchmarks}
\begin{tabular}{lcc}
\toprule
\textbf{Model} & \textbf{Accuracy [\%]} & \textbf{Avg. Latency [s]} \\
\midrule
\multicolumn{3}{c}{\textit{Online Models}} \\
\midrule
GPT-4o-2024-11-20 & 76.47 & 3.38 \\
GPT-4o-2024-08-06 & 74.12 & 4.16 \\
o3-mini-2025-01-31 & 85.88 & 21.08 \\
GPT-4-turbo-2024-04-09 & 81.18 & 15.73 \\
GPT-3.5-turbo-0125 & 57.65 & 5.12 \\
\midrule
\multicolumn{3}{c}{\textit{Locally Deployed Models}} \\
\midrule
\textbf{Llama3.1-8B} & \textbf{8.24} & \textbf{0.61} \\
\textbf{Llama3.2-3B} & \textbf{2.35} & \textbf{0.44} \\
Deepseek-R1-7B & 5.88 & 4.07 \\
Deepseek-R1-14B & 16.47 & 9.86 \\
Deepseek-R1-32B & 21.18 & 34.96 \\
Gemma2-9B & 8.24 & 0.82 \\
Gemma2-27B & 27.06 & 1.82 \\
Gemma3-12B & 18.82 & 1.14 \\
\textbf{Mistral-7B} & \textbf{27.06} & \textbf{0.64} \\
Mistral-small-22B & 16.47 & 1.31 \\
Mistral-small-24B & 24.71 & 1.15 \\
\bottomrule
\end{tabular}
\end{table}

\subsection{Dataset Preparation for Model Fine-Tuning} \label{subsec:dataset_preparation}

The small initial dataset used for benchmarking, described in Subsection~\ref{subsec:initial_benchmarking} was sufficient for preliminary evaluation. However, it was too small for effective fine-tuning of language models.
To address this limitation, we extended the dataset. The augmentation process was performed using the GPT-4 LLM, which demonstrated strong semantic understanding in the initial benchmarking experiments. It was prompted to use various augmentation techniques~\cite{wang2024comprehensive, gao-etal-2019-soft, karimi2021aeda, wei-zou-2019-eda}, including synonym replacement, paraphrasing, word reordering, and minor contextual additions. These techniques increased linguistic diversity without introducing new command semantics, enabling effective fine-tuning.
Beyond predefined commands, we integrated an out-of-scope (OOS) class to handle unrelated inputs. This enhances the training of the classifier to increase system robustness through more reliable rejection of irrelevant prompts.
The resulting extended dataset, comprising 17 command classes and 1,645 samples, provides increased linguistic coverage by incorporating a broader range of vocabulary and syntactic structures. It was partitioned into an 80/20 train-test split, providing a training set of 1,308 samples.

\subsection{Fine-Tuning and Model Optimization}
\label{subsec:fine_tuning}

The three models selected in Subsection~\ref{subsec:initial_benchmarking} for further fine-tuning, were: \texttt{Mistral 7B}, \texttt{LLaMA-3.1 8B}, and \texttt{LLaMA-3.2 3B}.

Training was implemented using the \textit{Unsloth}\footnote{\url{https://unsloth.ai}} framework to apply QLoRA, targeting all linear modules~\cite{hu2021loralowrankadaptationlarge}. Specifically, adaptations were applied to the attention projection layers ($q, k, v, o$)~\cite{vaswani2023attentionneed} and the MLP layers ($gate, up, down$)~\cite{shazeer2020gluvariantsimprovetransformer}.

To identify the optimal configuration, we conducted a grid search over the following discrete hyperparameter space:

\begin{itemize}
    \item \textbf{Learning Rates:} $1 \times 10^{-5}$ and $5 \times 10^{-5}$
    \item \textbf{Epochs:} 2 and 3
    \item \textbf{Gradient Accumulation Steps:} 4 and 8
    \item \textbf{LoRA Rank ($r$):} 16 and 32 (with $\alpha$ scaled accordingly)
\end{itemize}

Fine-tuning was performed in two stages.

\subsection{First Stage Fine-tuning}
In the first phase, we evaluated the receptiveness of all three models towards fine-tuning. Given the large number of hyperparameter combinations, and to reduce computational cost, the initial fine-tuning was conducted on a subset of five out of seventeen command classes extracted from the full training dataset ( Section \ref{subsec:dataset_preparation}). Accordingly, the evaluation was performed on the corresponding subset of the test dataset. All experiments in this phase were performed on an NVIDIA GeForce RTX 4090 GPU. The results are presented in Table~\ref{tab:best_model_configs_basic_commands} and identified \texttt{Mistral 7B} as the best-performing model, achieving the highest accuracy of 76.74\,\%.

\begin{table*}[htbb]
\centering
\caption{Initial Fine-Tuning for Basic Commands \\ Best Configurations for Selected Models}
\label{tab:best_model_configs_basic_commands}
\begin{tabular}{lcccccc}
\toprule
\textbf{Model} & \textbf{Accuracy [\%]} & \textbf{Avg. Inference [s]} & \textbf{Epochs} & \textbf{Grad Acc Steps} & \textbf{r} & \textbf{alpha} \\
\midrule
\multicolumn{7}{c}{\textit{Locally Deployed Models}} \\
\midrule
\textbf{Mistral-7B} & \textbf{76.74} & \textbf{0.92} & 2 & 4 & 32 & 32 \\
Llama3.1-8B  & 75.58 & 0.88 & 3 & 4 & 32 & 32 \\
Llama3.2-3B  & 73.26 & 0.67 & 3 & 4 & 32 & 32 \\
\bottomrule
\end{tabular}
\end{table*}

\subsection{Second Stage Fine-Tuning}
\texttt{Mistral 7B} was selected for full fine-tuning on the complete training dataset, using the best-performing hyperparameter configuration.
The full fine-tuning and evaluation was conducted on an NVIDIA GeForce RTX 3060 Laptop GPU using the full train and test data as described in Section \ref{subsec:dataset_preparation}.
The best-performing fine-tuned variant of Mistral-7B, configured as specified in Table~\ref{tab:best_model_params}, achieved a 97.63\,\% classification accuracy, which exceeded the performance of every evaluated cloud-based model (Table~\ref{tab:model_benchmarks}). Moreover, with an average inference latency of 1.39\,s, it delivered faster response times, demonstrating that locally fine-tuned open-weight models can surpass the online-based ones in both accuracy and efficiency.

\begin{table}[h]
\centering
\caption{Best model results and hyperparameter after second stage fine-tuning}
\label{tab:best_model_params}
\begin{tabular}{@{}ll@{}}
\toprule
\textbf{Parameter} & \textbf{Value} \\ \midrule
\textbf{Model Name} & \textbf{Mistral-7B-v0.3-bnb-4bit} \\
\textbf{Accuracy [\%]} & \textbf{97.63} \\
\textbf{Avg. Inference [s]} & \textbf{1.39} \\
Learning Rate & $5 \times 10^{-5}$ \\
Epochs & 2 \\
LoRA Rank ($r$) & 32 \\
LoRA Alpha ($\alpha$) & 32 \\
Gradient Accumulation Steps & 4 \\
Optimizer & AdamW 8-bit \\ \bottomrule
\end{tabular}
\end{table}

\section{DISCUSSION} \label{sec:discussion}
The experimental results demonstrate that, for narrowly scoped command classification tasks, domain-specific fine-tuning of lightweight local language models can outperform larger online-hosted systems. While cloud-based models such as GPT-4 variants provide strong zero-shot intent and contextual understanding, they introduce an inference burden in a constrained command-classification scenario. Additionally, cloud-based models introduce network dependency, making system performance sensitive to connectivity and external service conditions. In contrast, our locally fine-tuned Mistral-7B model operates entirely on-device and is optimized for domain-specific phrasing patterns, resulting in more consistent behavior under controlled inference settings.

Our findings further highlight the effectiveness of QLoRA-based fine-tuning for edge deployment scenarios. Fine-tuning 7B-scale models using 4-bit quantization enables high task-specific performance on resource-constrained hardware, such as a mobile laptop workstation equipped with an NVIDIA GeForce RTX 3060 GPU. Our fine-tuning approach improves the accuracy of the final candidate model from 27.06\,\% to 97.64\,\%.

Despite these strengths, several limitations remain. First, although the dataset was expanded through GPT-4-based augmentation to increase linguistic diversity, it remains partially synthetic and text-based. While the augmentation process introduced paraphrastic variation, it may not capture the full distribution of possible inputs, including incomplete sentences, hesitations, or unconventional formulations. Consequently, future work should incorporate real-world operator transcripts collected under authentic racing conditions to better reflect natural language usage and improve generalization.

Moreover, the current system operates within a strictly predefined command space. Although this constraint enhances safety and interpretability, it limits conversational flexibility. Extending the system toward more adaptive dialogue capabilities would require hierarchical command validation and formally defined safety constraints to prevent unintended behavior.

\section{CONCLUSION \& OUTLOOK}
In this work, we presented \textit{Jarvis}, an offline voice assistant architecture for high-level control of autonomous racing vehicles. By integrating wake-word detection, local speech-to-text transcription, and a fine-tuned lightweight language model for intent classification, we developed a system with quick response times and network-independent operation, suitable for use as a voice assistant in high-speed racing applications.

The presented architecture provides an open-source framework for an interactive voice assistant using speech-to-command classification. Beyond racing, the approach is transferable to other robotic systems where deterministic performance, low latency, and network independence are essential.

Future work includes multi-turn dialogue and bidirectional telemetry querying.

\bibliographystyle{IEEEtran}
\bibliography{Literatur_Frederik_Werner_TUM_File, Literatur_Simon_Sagmeister_TUM, Literatur_Daniel_Henel_TUM}

\end{document}